\documentclass[11pt,a4paper]{article}
\usepackage[hyperref]{acl2020}
\usepackage{times}
\usepackage{latexsym}

\usepackage{microtype}

\aclfinalcopy 

\usepackage{graphicx}
\usepackage{booktabs}
\usepackage{float}
\usepackage{xcolor}
\usepackage{xspace}
\usepackage{url}
\usepackage{changepage}
\usepackage{bold-extra}

\newcommand{\cyan}[1]{\textcolor{cyan}{#1}}
\newcommand{\magenta}[1]{\textcolor{magenta}{#1}}
\newcommand{\orange}[1]{\textcolor{orange}{#1}}

\newcommand\bioread{\textsc{bioread}\xspace}

\newcommand\biomrc{\textsc{biomrc}\xspace}
\newcommand\biocloze\biomrc

\newcommand\biolarge{\textsc{large}\xspace}
\newcommand\biolite{\textsc{lite}\xspace}
\newcommand\biotiny{\textsc{tiny}\xspace}
\newcommand\metamap{\textsc{metamap}\xspace}
\newcommand\mrc{\textsc{mrc}\xspace}
\newcommand\mlp{\textsc{mlp}\xspace}
\newcommand\aoareader{\textsc{aoa-reader}\xspace}
\newcommand\asreader{\textsc{as-reader}\xspace}
\newcommand\baseone{\textsc{base1}\xspace}
\newcommand\basetwo{\textsc{base2}\xspace}
\newcommand\basethree{\textsc{base3}\xspace}
\newcommand\basethreeplus{\textsc{base3+}\xspace}
\newcommand\basefour{\textsc{base4}\xspace}
\newcommand\pubtator{\textsc{pubtator}\xspace}
\newcommand\pubmed{\textsc{pubmed}\xspace}
\newcommand\pmc{\textsc{pubmed central}\xspace}

\newcommand\cnndaily{\textsc{cnn} and Daily Mail\xspace}
\newcommand\bert{\textsc{bert}\xspace}

\newcommand\nltk{\textsc{nltk}\xspace}
\newcommand\squad{\textsc{squad}\xspace}

\newcommand\bioasq{\textsc{bioasq}\xspace}
\newcommand\clicr{\textsc{clicr}\xspace}
\newcommand\qa{\textsc{qa}\xspace}
\newcommand\dnorm{\textsc{dnorm}\xspace}
\newcommand\scibert{\textsc{scibert}\xspace}

\newcommand\scibertreadersum{\textsc{scibert-sum-reader}\xspace}
\newcommand\scibertsumreader\scibertreadersum
\newcommand\scibertmaxreader\scibertreadermax 
\newcommand\scibertreadermax{\textsc{scibert-max-reader}\xspace}
\newcommand\bertreaderf{\textsc{scibert-sum-reader}\xspace}
\newcommand\bertreaderfmax{\textsc{scibert-max-reader}\xspace}
\newcommand\septok{[\textsc{sep}]\xspace}
\newcommand\masktok{[\textsc{mask}]\xspace}

\title{\normalfont{\textbf{
\biomrc: A Dataset for Biomedical Machine Reading Comprehension
}}}

\author{\parbox{12cm}{\centering Petros Stavropoulos\textsuperscript{1,2}, Dimitris Pappas\textsuperscript{1,2}, Ion Androutsopoulos\textsuperscript{1}, \vspace{0.5mm} \\
Ryan McDonald\textsuperscript{3,1} \vspace{1.5mm}} \\
  \textsuperscript{1}Department of Informatics, Athens University of Economics and Business, Greece \\
  \textsuperscript{2}Institute for Language and Speech Processing, Research Center `Athena', Greece\\
  \textsuperscript{3}Google Research\\
  {\parbox{12cm}{\centering \tt \{pstav1993, pappasd, ion\}@aueb.gr  ryanmcd@google.com}} \\}

\date{}

\begin{document}
\maketitle
\begin{abstract}
We introduce \biomrc, a large-scale cloze-style biomedical \mrc dataset.
Care was taken to reduce noise, compared to the previous \bioread dataset of \newcite{Pappas2018BioReadAN}. 
Experiments show that simple heuristics do not perform well on the new dataset, and that two neural \mrc models that had been tested on \bioread perform much better on \biomrc, indicating that the new dataset is indeed less noisy or at least that its task is more feasible. 
Non-expert human performance is also higher on the new dataset compared to \bioread, and biomedical experts perform even better. 
We also introduce a new \bert-based \mrc model, the best version of which substantially outperforms all other methods tested, reaching or surpassing the accuracy of biomedical experts in some experiments. 
We make the new dataset available in three different sizes, also releasing our code, and providing a leaderboard. 
\end{abstract}

\section{Introduction}

Creating large corpora with human annotations is a demanding process in both time and resources.
Research teams often turn to distantly supervised or unsupervised methods to extract training examples from textual data.
In machine reading comprehension (\mrc) \cite{Moritz_Hermann_et_al_cnndaily}, a training instance can be automatically constructed by taking an unlabeled passage of multiple sentences, along with another smaller part of text, also unlabeled, usually the next sentence. Then a named entity of the smaller text is replaced by a placeholder. In this setting, \mrc systems are trained (and evaluated for their ability) to read the passage and the smaller text, and guess the named entity that was replaced by the placeholder, which is typically one of the named entities of the passage. This kind of question answering (\qa) is also known as cloze-type questions \cite{Taylor1953ClozePA}.
Several datasets have been created following this approach either using books \cite{Hill_et_al_CBTest, Bajgar_et_al_BookTest} or news articles \cite{Moritz_Hermann_et_al_cnndaily}.
Datasets of this kind are noisier than \mrc datasets containing human-authored questions and manually annotated passage spans that answer them \cite{Rajpurkar2016SQuAD10, rajpurkar_etal_2018_squadv2, Nguyen_et_al_msmarco}. They require no human annotations, however, which is particularly important in biomedical question answering, where employing annotators with appropriate expertise is costly. For example, the \bioasq \qa dataset \cite{Tsatsaronis_et_al_bioasq} currently contains approximately 3k questions, much fewer than the 100k questions of a \squad \cite{Rajpurkar2016SQuAD10}, exactly because it relies on expert annotators. 

To bypass the need for expert annotators and produce a biomedical \mrc dataset large enough to train (or pre-train) deep learning models, \newcite{Pappas2018BioReadAN} adopted the cloze-style questions approach. They used the full text of unlabeled biomedical articles from \pmc,\footnote{\url{https://www.ncbi.nlm.nih.gov/pmc/}} and \metamap \cite{Aronson2010AnOO_metamap} to annotate the biomedical entities of the articles.
They extracted sequences of $21$ sentences from the articles.
The first $20$ sentences were used as a passage and the last sentence as a cloze-style question.
A biomedical entity of the `question' was replaced by a placeholder, and systems have to guess which biomedical entity of the passage can best fill the placeholder. 
This allowed Pappas et al.\ to produce a dataset, called \bioread, of approximately 16.4 million questions. As the same authors reported, however, the mean accuracy of three humans on a sample of 30 questions from \bioread was only 68\%. Although this low score may be due to the fact that the three subjects were not biomedical experts, it is easy to see, by examining samples of \bioread, that many examples of the dataset do not make sense.
Many instances contain passages or questions crossing article sections, or originating from the references sections of articles, or they include captions and footnotes (Table~\ref{tab:BioReadExamples}). Another source of noise is \metamap, which often misses or mistakenly identifies biomedical entities (e.g., it often annotates `to' as the country Togo). 

\begin{table}[]
    \centering
    \small
    \begin{tabular}{c}
    \hline
        \textbf{`question' originating from caption}: \\
        ``figure 4 htert
        @entity6 and @entity4 \textsc{xxxx} cell invasion.''\\
    \hline
        \textbf{`question' originating from reference}:\\
        ``2004 , 17 , 250 257
        .14967013 c samuni y. ; samuni u. ; \\
        goldstein
        s. the use of cyclic \textsc{xxxx} as hno scavengers .''\\
    \hline
        \textbf{`passage' containing captions}: \\
        ``figure 2: distal \textsc{unk} showing high insertion 
        of rectum \\
        into common channel. figure 3: 
        illustration of the cloacal \\ 
        malformation. 
        figure 4: @entity5 showing \textsc{unk}''\\
    \hline
    \end{tabular}
    \vspace*{-2mm}
    \caption{
    Examples of noisy \bioread data. \textsc{xxxx} is the placeholder, and \textsc{unk} is the `unknown' token.
    }
    \vspace*{-5mm}
    \label{tab:BioReadExamples}
\end{table}

In this paper, we introduce \biomrc, a new dataset for biomedical \mrc that can be viewed as an improved version of \bioread. To avoid crossing sections, extracting text from references, captions, tables etc., we use abstracts and titles of biomedical articles as passages and questions, respectively, which are clearly marked up in \pubmed data, instead of using the full text of the articles.
Using titles and abstracts is a decision that favors precision over recall.
Titles are likely to be related to their abstracts,
which reduces the noise-to-signal ratio significantly and  
makes it less likely to generate irrelevant questions for a passage.
We replace a biomedical entity in each title with a placeholder, and we require systems to guess the hidden entity by considering the entities of the abstract as candidate answers. Unlike \bioread, we use \pubtator \cite{PubTator2012}, a repository that provides approximately $25$ million abstracts and their corresponding titles from \pubmed, with multiple annotations.\footnote{Like \pubmed, \pubtator is supported by \textsc{ncbi}. Consult: \url{www.ncbi.nlm.nih.gov/research/pubtator/}} We use \dnorm's biomedical entity annotations, which are more accurate than \metamap's \cite{Leaman_et_al_pubtator_dnorm_vs_metamap}. We also perform several checks, discussed below, to discard passage-question instances that are too easy, and we show that the accuracy of experts and non-expert humans reaches 85\% and 82\%, respectively, on a sample of 30 instances for each annotator type, which is an indication that the new dataset is indeed less noisy, or at least that the task is more feasible for humans. Following \newcite{Pappas2018BioReadAN}, we release two versions of \biomrc, \biolarge and \biolite, containing 812k and 100k instances respectively, for researchers with more or fewer resources, along with the 60 instances (\biotiny) humans answered. Random samples from \biomrc \biolarge where selected to create \biolite and \biotiny. \biomrc \biotiny is used only as a test set; it has no training and validation subsets.

We tested on \biomrc \biolite the two deep learning \mrc models that \newcite{Pappas2018BioReadAN} had tested on \bioread \biolite, namely Attention Sum Reader (\asreader) \cite{Kadlec2016TextUW} and Attention Over Attention Reader (\aoareader) \cite{Cui2017AttentionoverAttentionNN}.
Experimental results show that \asreader and \aoareader perform better on \biomrc, with the accuracy of \aoareader reaching 70\% compared to the corresponding 52\% accuracy of \newcite{Pappas2018BioReadAN}, which is a further indication that the new dataset is less noisy or that at least its task is more feasible.
We also developed a new \bert-based \cite{Devlin_et_al_BERT} \mrc model, the best version of which (\scibertmaxreader) performs even better, with its accuracy reaching 80\%.
We encourage further research on biomedical \mrc by making our code and data publicly available, and by creating an on-line leaderboard for \biomrc.\footnote{Our code, data, and information about the leaderboard will be available at  \url{http://nlp.cs.aueb.gr/publications.html}.} 

\begin{figure*}[t]
\resizebox{\textwidth}{!}{
\small
\begin{tabular}{|c|l|}
\hline
Passage & \begin{tabular}[c]{@{}p{15.2cm}@{}}BACKGROUND: Most brain metastases arise from \orange{\textbf{@entity0}} . Few studies compare the brain regions they involve, their numbers and intrinsic attributes. METHODS: Records of all \cyan{@entity1} referred to Radiation Oncology for treatment of symptomatic brain metastases were obtained. Computed tomography (n = 56) or magnetic resonance imaging (n = 72) brain scans were reviewed. RESULTS: Data from 68 breast and 62 \cyan{@entity2} \cyan{@entity1} were compared. Brain metastases presented earlier in the course of the lung than of the \orange{\textbf{@entity0}} \cyan{@entity1} (p = 0.001). There were more metastases in the cerebral hemispheres of the breast than of the \cyan{@entity2} \cyan{@entity1} (p = 0.014). More \orange{\textbf{@entity0}} \cyan{@entity1} had cerebellar metastases (p = 0.001). The number of cerebral hemisphere metastases and presence of cerebellar metastases were positively correlated (p = 0.001). The prevalence of at least one \cyan{@entity3} surrounded with $>2$ cm of \cyan{@entity4} was greater for the lung than for the breast \cyan{@entity1} (p = 0.019). The \cyan{@entity5} type, rather than the scanning method, correlated with differences between these variables. CONCLUSIONS: Brain metastases from lung occur earlier, are more \cyan{@entity4} , but fewer in number than those from \orange{\textbf{@entity0}} . Cerebellar brain metastases are more frequent in \orange{\textbf{@entity0}} .\end{tabular} \\
\hline
Candidates & 
\begin{tabular}[c]{@{}p{15.2cm}@{}}
\orange{\textbf{@entity0}} : {[}`breast and lung cancer'{]} ; 
\cyan{@entity1} : {[}`patients'{]} ; 
\cyan{@entity2} : {[}`lung cancer'{]} ; \\
\cyan{@entity3} : {[}`metastasis'{]} ; 
\cyan{@entity4} : {[}`edematous', `edema'{]} ; 
\cyan{@entity5} : {[}`primary tumor'{]}
\end{tabular}\\ \hline
Question    & 
\begin{tabular}[c]{@{}p{15.2cm}@{}}Attributes of brain metastases from \magenta{XXXX} .\end{tabular}
\\ \hline
Answer   & 
\begin{tabular}[c]{@{}p{6.2cm}@{}}\orange{\textbf{@entity0}} : {[}`breast and lung cancer'{]}\end{tabular}\\ \hline
\end{tabular}
}
\caption{Example passage-question instance of \biomrc. The passage is the abstract of an article, with biomedical entities replaced by @entity$N$ pseudo-identifiers. The original entity names are shown in square brackets. Both `edematous' and `edema' are replaced by `@entity4', because \pubtator considers them synonyms. The question is the title of the article, with a biomedical entity replaced by \textsc{xxxx}. @entity0 is the correct answer. 
}
\label{fig:fig1}
\label{fig:BioMRCExample}
\end{figure*}

\section{Dataset Construction}

\begin{table*}[t]
\begin{center}
\resizebox{\textwidth}{!}{
\begin{tabular}{c|cccc|cccc|ccc}
\hline
&
\multicolumn{4}{|c|}{\textbf{\biomrc \biolarge}}
&
\multicolumn{4}{|c|}{\textbf{\biomrc \biolite}}
&
\multicolumn{3}{|c}{\textbf{\biomrc \biotiny}}
\\
\hline
                           & \textbf{Training} & \textbf{Development} & \textbf{Test} & \textbf{Total} & \textbf{Training} & \textbf{Development} & \textbf{Test} & \textbf{Total} & \textbf{Setting A} & \textbf{Setting B} & \textbf{Total} \\ \hline \hline
\textbf{Instances}         & 700,000            & 50,000                & 62,707         & 812,707 & 87,500             & 6,250                 & 6,250          & 100,000 & 30                 & 30          & 60         \\
\textbf{Avg candidates}    & 6.73              & 6.68                 & 6.68          & 6.72          & 6.72              & 6.68                 & 6.65          & 6.71 & 6.60                 & 6.57          & 6.58 \\
\textbf{Max candidates}    & 20                & 20                   & 20            & 20           & 20                & 20                   & 20            & 20 & 13                   & 11            & 13 \\
\textbf{Min candidates}    & 2                 & 2                    & 2             & 2             & 2                 & 2                    & 2             & 2 & 2                    & 3             & 2 \\
\textbf{Avg abstract len.}  & 253.79            & 257.41               & 253.70        & 254.01     & 253.78            & 257.32               & 255.56        & 254.11  & 248.13               & 264.37        & 256.25  \\
\textbf{Max abstract len.}  & 543               & 516                  & 511           & 543        & 519               & 500                  & 510           & 519   & 371                  & 386           & 386  \\
\textbf{Min abstract len.}  & 57                & 89                   & 77            & 57         & 60                & 109                  & 103           & 60   & 147                  & 154           & 147 \\
\textbf{Avg title len.} & 13.93             & 14.28                & 13.99         & 13.96      & 13.89             & 14.22                & 14.09         & 13.92  & 14.17                & 14.70         & 14.43  \\
\textbf{Max title len.} & 51                & 46                   & 43            & 51             & 49                & 40                   & 42            & 49 & 21                   & 35            & 35\\
\textbf{Min title len.} & 3                 & 3                    & 3             & 3              & 3                 & 3                    & 3             & 3 & 6                    & 4             & 4 \\ \hline
\end{tabular}
}
\caption{Statistics of \biomrc \biolarge, \biolite, \biotiny. The questions of the \biotiny version were answered by humans. All lengths are measured in tokens using a whitespace tokenizer.
}
\label{tab:tab1}
\label{tab:statistics}
\end{center}
\end{table*}

Using \pubtator, we gathered approx.\ $25$ million  abstracts and their titles.
We discarded articles with titles shorter than 15 characters or longer than 60 tokens, articles without abstracts, or with abstracts shorter than 100 characters, or fewer than 10 sentences. We also removed articles with abstracts containing fewer than 5 entity annotations, or fewer than 2 or more than 20 distinct biomedical entity identifiers. (\pubtator assigns the same identifier to all the synonyms of a biomedical entity; e.g., `hemorrhagic stroke' and `stroke' have the same identifier `\textsc{mesh:d020521}'.) We also discarded articles containing entities not linked to any of the ontologies used by \pubtator,\footnote{\pubtator uses the Open Biological and Biomedical Ontology (\textsc{obo}) Foundry, which comprises over 60 ontologies.} or entities linked to multiple ontologies (entities with multiple ids), or entities whose spans overlapped with those of other entities. 
We also removed articles with no entities in their titles, 
and articles with no entities shared by the title and abstract.\footnote{A further reason for using the title as the question is that the entities of the titles are typically mentioned in the abstract.} 
Finally, to avoid making the dataset too easy for a system that would always select the entity with the most occurrences in the abstract, we removed a passage-question instance if the most frequent entity of its passage (abstract) was also the answer to the cloze-style question (title with placeholder); 
if multiple entities had the same top frequency in the passage, the instance was retained.
We ended up with approx.\ 812k passage-question instances, which form \biomrc \biolarge, split into training, development, and test subsets (Table~\ref{tab:statistics}). The \biolite and \biotiny versions of \biomrc are subsets of \biolarge.

In all versions of \biomrc (\biolarge, \biolite, \biotiny), the entity identifiers of \pubtator are replaced by pseudo-identifiers of the form @entity$N$ (Fig.~\ref{fig:fig1}), as in the \cnndaily datasets \cite{Moritz_Hermann_et_al_cnndaily}. We provide all \biomrc versions in two forms, corresponding to what \newcite{Pappas2018BioReadAN} call Settings A and B in \bioread.\footnote{\newcite{Pappas2018BioReadAN} actually call `option~a' and `option~b' our Setting B and A, respectively.} 
In Setting A, each pseudo-identifier has a global scope, meaning that each biomedical entity has a unique pseudo-identifier in the whole dataset.
This allows a system to learn information about the entity represented by a pseudo-identifier from all the occurrences of the pseudo-identifier in the training set.
For example after seeing the same pseudo-identifier multiple times a model may learn that it stands for a drug, or that a particular pseudo-identifier tends to neighbor with specific words. Then, much like a language model, a system may guess the pseudo-identifier that should fill in the placeholder even without the passage, or at least it may infer a prior probability for each candidate answer.
In contrast, Setting B uses a local scope,  
i.e., it restarts the numbering of the pseudo-identifiers (from @entity0) anew in each passage-question instance. This forces the models to rely only on 
information about the entities that can be inferred from the particular passage and question. 
This corresponds to a non-expert answering the question,
who does not have any prior knowledge of the biomedical entities.

Table~\ref{tab:statistics} provides statistics on \biomrc. In \biotiny, we use 30 different passage-question instances in Settings A and B, because in both settings we asked the same humans to answer the questions, and we did not want them to remember instances from one setting to the other. In \biolarge and \biolite, the instances are the same across the two settings, apart from the numbering of the entity identifiers.

\section{Experiments and Results}
\label{sec:exp_results}

We experimented only on \biomrc \biolite and \biotiny, since we did not have the computational resources to train the neural models we considered on the \biolarge version of \bioread. \newcite{Pappas2018BioReadAN} also reported experimental results only on a \biolite version of their \bioread dataset. We hope that others may be able to experiment on \biomrc \biolarge, and we make our code 
available, as already noted.

\subsection{Methods}
We experimented with the four basic baselines (\baseone--4) that \newcite{Pappas2018BioReadAN} used in \bioread, the two neural \mrc models used by the same authors, \asreader \cite{Kadlec2016TextUW} and \aoareader \cite{Cui2017AttentionoverAttentionNN}, and a \bert-based  \cite{Devlin_et_al_BERT} model we developed. 

\begin{figure}[tb]
\centering

\includegraphics[width=3.0in,height=1.5in]{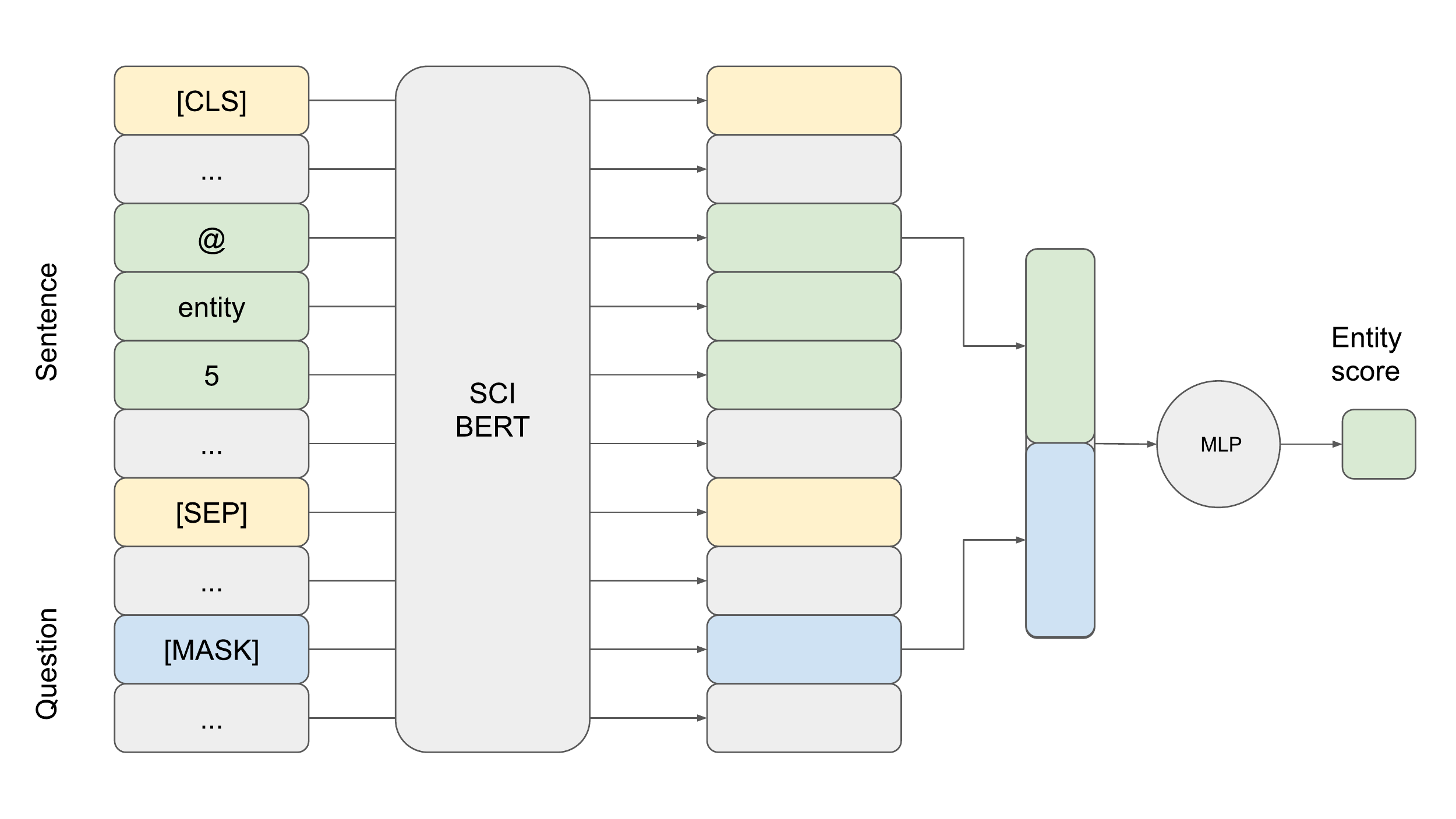}
\hspace*{-5mm}
\caption{
Illustration of our \scibert-based models. Each sentence of the passage is concatenated with the question and fed to \scibert. The top-level embedding produced by \scibert for the first sub-token of each candidate answer is concatenated with the top-level embedding of \masktok (which replaces the placeholder \textsc{xxxx}) of the question, and they are fed to an \mlp, which produces the score of the candidate answer. In \scibertsumreader, the scores of multiple occurrences of the same candidate are summed, whereas \scibertmaxreader takes their maximum. 
}
\vspace*{-4mm}
\label{fig:scibert_reader_backbone}
\end{figure}

\smallskip
\noindent\textbf{Basic baselines}: \baseone, 2, 3 return the first, last, and the entity that occurs most frequently in the passage (or randomly one of the entities with the same highest frequency, if multiple exist), respectively. 
Since in \bioread the correct answer is never (by construction) the most frequent entity of the passage, unless there are multiple entities with the same highest frequency, \basethree performs poorly. Hence, we also include a variant, \basethreeplus, which randomly selects one of the entities of the passage with the same highest frequency, if multiple exist, otherwise it selects the entity with the second highest frequency. 
\basefour extracts all the token $n$-grams from the passage that include an entity identifier (@entity$N$), and all the $n$-grams from the question that include the placeholder (\textsc{xxxx}).\footnote{We tried $n=2,\dots,6$ and use $n=3$, which gave the best accuracy on the development set of \biomrc \biolarge.}
Then for each candidate answer (entity identifier), it counts the tokens shared between the $n$-grams that include the candidate and the $n$-grams that include the placeholder.
The candidate with the most shared tokens is selected.
These baselines are used to check that the questions cannot be answered by simplistic heuristics \cite{Chen_et_al_vs_cnndailymail}.

\smallskip
\noindent\textbf{Neural baselines}: We use the same implementations of \asreader \cite{Kadlec2016TextUW} and \aoareader \cite{Cui2017AttentionoverAttentionNN} as \newcite{Pappas2018BioReadAN}, who also provide short descriptions of these neural models, not provided here to save space. The hyper-parameters of both methods were tuned on the development set of \biomrc \biolite.

\begin{table*}[ht]
\begin{center}
\resizebox{\textwidth}{!}{
\begin{tabular}{c|ccccccc|ccccccc}
\hline
 &
\multicolumn{7}{|c|}{\textbf{\biomrc Lite -- Setting A}} &
\multicolumn{7}{|c}{\textbf{\biomrc Lite -- Setting B}} 
\\
\hline
& \textbf{Train} & \textbf{Dev} & \textbf{Test} & \textbf{Train} & \textbf{All} & \textbf{Word} & \textbf{Entity} & \textbf{Train} & \textbf{Dev} & \textbf{Test} & \textbf{Train} & \textbf{All} & \textbf{Word} & \textbf{Entity} \\
\textbf{Method}    & \textbf{Acc} & \textbf{Acc} & \textbf{Acc} & \textbf{Time} & \textbf{Params} & \textbf{Embeds} & \textbf{Embeds} & \textbf{Acc} & \textbf{Acc} & \textbf{Acc} & \textbf{Time} & \textbf{Params} & \textbf{Embeds} & \textbf{Embeds} \\
\hline
\hline
\textbf{\baseone} 
& 37.58 & 36.38 & 37.63 & 0 & 0 & 0 & 0
& 37.58 & 36.38 & 37.63 & 0 & 0 & 0 & 0\\
\textbf{\basetwo} 
& 22.50 & 23.10 & 21.73 & 0 & 0 & 0 & 0
& 22.50 & 23.10 & 21.73 & 0 & 0 & 0 & 0\\
\textbf{\basethree} 
& 10.03 & 10.02 & 10.53 & 0 & 0 & 0 & 0
& 10.03 & 10.02 & 10.53 & 0 & 0 & 0 & 0\\
\textbf{\basethreeplus} 
& 44.05 & 43.28 & 44.29 & 0 & 0 & 0 & 0
& 44.05 & 43.28 & 44.29 & 0 & 0 & 0 & 0\\
\textbf{\basefour}
& 56.48 & 57.36 & 56.50 & 0 & 0 & 0 & 0
& 56.48 & 57.36 & 56.50 & 0 & 0 & 0 & 0\\
\hline
\textbf{\asreader}
& \textbf{84.63} & 62.29 & 62.38 & 18 x 0.92 hr & 12.87M & 12.69M & 1.59M
& 79.64 & 66.19 & 66.19 & 18 x 0.65 hr & 6.82M & 6.66M & 0.60k\\
\textbf{\aoareader}
& 82.51 & 70.00 & 69.87 & 29 x 2.10 hr & 12.87M & 12.69M & 1.59M
& \textbf{84.62} & 71.63 & 71.57 & 36 x 1.82 hr & 6.82M & 6.66M & 0.60k\\
\textbf{\bertreaderf}
& 71.74 & 71.73 & 71.28 & 11 x 4.38 hr & \textbf{154k} & \textbf{0} & \textbf{0}
& 68.92 & 68.64 & 68.24 & 6 x 4.38 hr & \textbf{154k} & \textbf{0} & \textbf{0}\\
\textbf{\bertreaderfmax}
& 81.38 & \textbf{80.06} & \textbf{79.97} & 19 x 4.38 hr & \textbf{154k} & \textbf{0} & \textbf{0}
& 81.43 & \textbf{80.21} & \textbf{79.10} & 15 x 4.38 hr & \textbf{154k} & \textbf{0} & \textbf{0}\\
\hline
\end{tabular}
}
\vspace*{-2mm}
\caption{
Training, development, test accuracy (\%) on \biomrc \biolite in Settings A (global scope of entity identifiers) and B (local scope), training times (epochs $\times$ time per epoch), and number of trainable parameters (total, word embedding parameters, entity identifier embedding parameters). In the lower zone (neural methods), the difference from each accuracy score to the next best is statistically significant ($p < 0.02$). We used singe-tailed Approximate Randomization \cite{P18-1128}, randomly swapping the 
answers to 50\% of the questions for 10k iterations.
}
\vspace*{-5mm}
\label{tab:tab3}
\end{center}
\end{table*}

\smallskip
\noindent\textbf{\bert-based model}:
We use \scibert \cite{beltagy-etal-2019-scibert}, a pre-trained \bert \cite{Devlin_et_al_BERT} model for scientific text. 
\scibert is pre-trained on 1.14 million articles from Semantic Scholar,\footnote{\url{https://www.semanticscholar.org/}}
of which 82\% (935k) are biomedical and the rest come from  computer science.
For each passage-question instance, we split the passage into sentences using \nltk \cite{Loper2002NLTKTN}. For each sentence,
we concatenate it (using \bert's \septok token) with the question, after replacing the \textsc{xxxx} with \bert's \masktok token, and we feed the concatenation to \scibert (Fig.~\ref{fig:scibert_reader_backbone}). We collect \scibert's top-level vector representations of the entity identifiers (@entity$N$) of the sentence and \masktok.\footnote{\bert's tokenizer splits the entity identifiers into sub-tokens \cite{Devlin_et_al_BERT}.
We use the first one.
The top-level token representations of \bert are context-aware, and it is common to use the first or last sub-token of each named-entity.}
For each entity of the sentence, we concatenate its top-level representation with that of \masktok, and we feed them to a Multi-Layer Perceptron (\mlp) to obtain a score for the particular entity (candidate answer). We thus obtain a score for all the entities of the passage.
If an entity occurs multiple times in the passage, we take the sum or the maximum of the scores of its occurrences.
In both cases, a softmax is then applied to the scores of all the entities, and the entity with the maximum score is selected as the answer. We call this model \scibertsumreader or \scibertmaxreader, depending on how it aggregates the scores of multiple occurrences of the same entity.

\scibertsumreader is closer to \asreader and \aoareader, which also sum the scores of multiple occurrences of the same entity. This summing aggregation, however, favors entities with several occurrences in the passage, even if the scores of all the occurrences are low. Our experiments indicate that \scibertmaxreader performs better. In all cases, we only update the parameters of the \mlp during training, keeping the parameters of \scibert frozen to their pre-trained values to speed up training. With more computing resources, it may be possible to improve the scores of \scibertmaxreader (and \scibertsumreader) further by fine-tuning \scibert on \biomrc training data.

\subsection{Results on \biomrc \biolite}

Table~\ref{tab:tab3} reports the accuracy of all methods on \biomrc \biolite for Settings A and B. In both settings, all the neural models clearly outperform all the basic baselines, with \basethree (most frequent entity of the passage) performing worst and \basethreeplus performing much better, as expected. In both settings, \scibertmaxreader clearly outperforms all the other methods on both the development and test sets. The performance of \scibertsumreader is approximately ten percentage points worse than \scibertmaxreader's on the development and test sets of both settings, indicating that the superior results of \scibertmaxreader are to a large extent due to the different aggregation function (max instead of sum) it uses to combine the scores of multiple occurrences of a candidate answer, not to the extensive pre-training of \scibert.
\aoareader, which does not employ any pre-training, is competitive to \scibertsumreader in Setting A, and performs better than \scibertsumreader in Setting B, which again casts doubts on the value of \scibert's extensive pre-training.
We expect, however, that the performance of the \scibert-based models, could be improved further by fine-tuning \scibert's parameters. 

The performance of \scibertsumreader is slightly better in Setting A than in Setting B, which might suggest that the model manages to capture global properties of the entity pseudo-identifiers from the entire training set. However, the performance of \scibertmaxreader is almost the same across the two settings, which contradicts the previous hypothesis. Furthermore, the development and test performance of \asreader and \aoareader is higher in Setting B than A, indicating that these two models do not capture global properties of entities well, performing better when forced to consider only the information of the particular passage-question instance. Overall, we see no strong evidence that the models we considered are able to learn global properties of the entities.

In both Settings A and B, \aoareader performs better than \asreader, which was expected since it uses a more elaborate attention mechanism, at the expense of taking longer to train (Table~\ref{tab:tab3}).\footnote{We trained all models for a maximum of $40$ epochs, using early stopping on the dev.\ set, with patience of $3$ epochs.} The two \scibert-based models are also competitive in terms of training time, because we only train the \mlp (154k parameters) on top of \scibert, keeping the parameters of \scibert frozen. 

\begin{figure*}[ht] 
\centering
\includegraphics[width=\textwidth,height=3in]{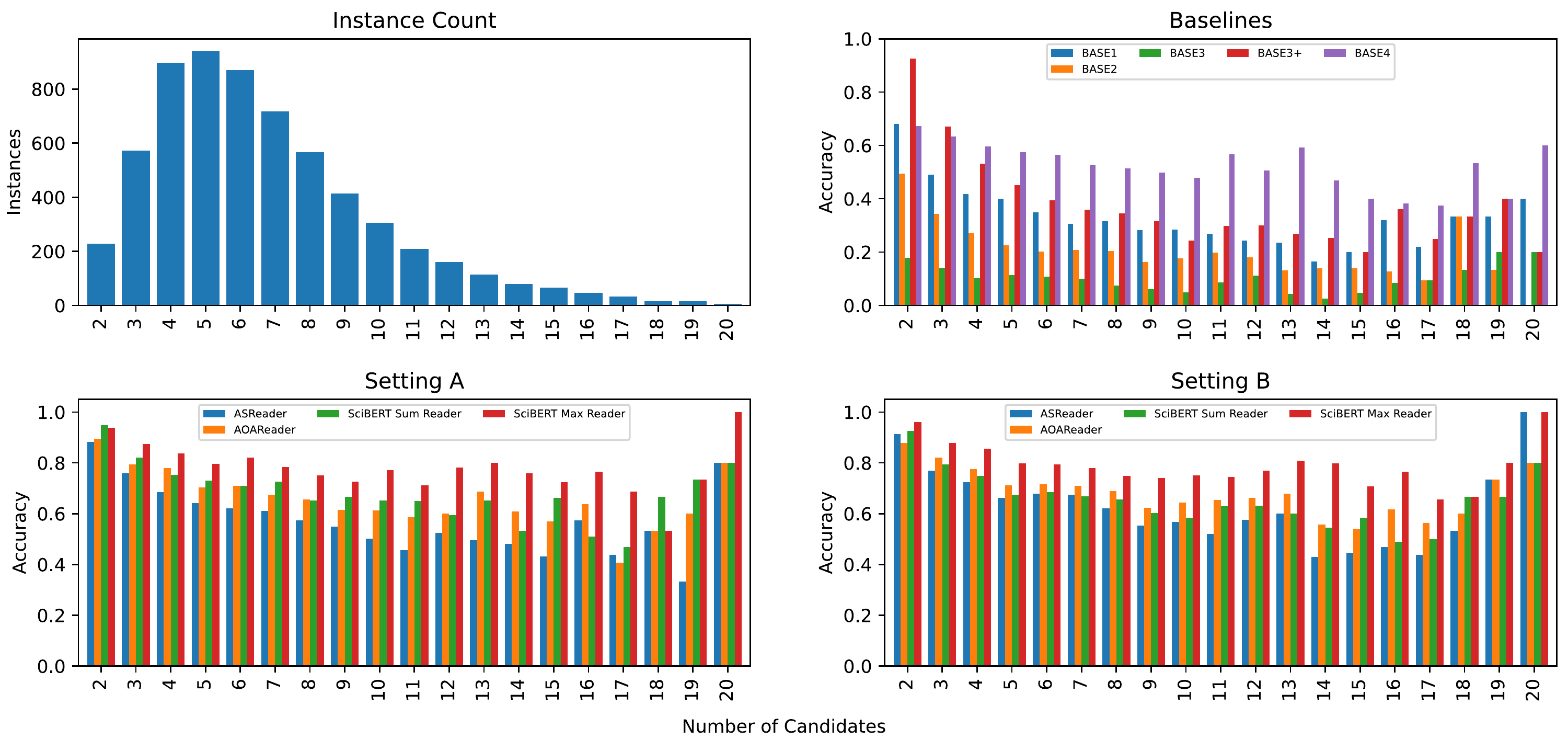}
\vspace*{-3mm}
\caption{More detailed statistics and results on the development subset of \biomrc \biolite. Number of passage-question instances with 2, 3, \dots, 20 candidate answers (top left). Accuracy (\%) of the basic baselines (top right). Accuracy (\%) of the neural models in Settings A (bottom left) and B (bottom right).
}
\label{fig:results_and_stats}
\end{figure*}

The trainable parameters of \asreader and \aoareader are almost double in Setting A compared to Setting B. To some extent, this difference is due to the fact that for both models we learn a word embedding for each @entity$N$ pseudo-identifier, and in Setting A the numbering of the identifiers is not reset for each passage-question instance, leading to many more  pseudo-identifiers (31.77k pseudo-identifiers in the vocabulary of Setting A vs.\ only 20 in Setting B); this accounts for a difference of 1.59M parameters.\footnote{Hyper-parameter tuning led to 50- and 30-dimensional word embeddings in Settings A, B, respectively. \asreader and \aoareader learn word embeddings from the training set, without using pre-trained embeddings.}
The rest of the difference in total parameters (from Setting A to B) is due to the fact that we tuned the hyper-parameters of each model separately for each setting (A, B), on the corresponding development set. Hyper-parameter tuning was performed separately for each model in each setting, but led to the same numbers of trainable parameters for \asreader and \aoareader, because the trainable parameters are dominated by the parameters of the word embeddings.
Note that the hyper-parameters of the two \scibert-based models (of their \mlp{s}) were very minimally tuned, hence these models may perform even better with more extensive tuning. 

\aoareader was also better than \asreader in the experiments of \newcite{Pappas2018BioReadAN} on a \biolite version of their \bioread dataset, but the development and test accuracy of \aoareader in Setting A of \bioread was reported to be only 52.41\% and 51.19\%, respectively (cf.\ Table~\ref{tab:tab3}); in Setting B, it was 50.44\% and 49.94\%, respectively. The much higher scores of \aoareader (and \asreader) on \biomrc \biolite are an indication that the new dataset is less noisy, or that the task is at least more feasible for machines. The results of \newcite{Pappas2018BioReadAN} were slightly higher in Setting A than in Setting B, suggesting that \aoareader was able to benefit from the global scope of entity identifiers, unlike our findings in \biomrc.\footnote{For \asreader, \newcite{Pappas2018BioReadAN} report results only for Setting B: 37.90\% development and 42.01\% test accuracy on \bioread \biolite. They did not  consider \bert-based models.}

Figure~\ref{fig:results_and_stats} shows how many passage-question instances of the development subset of \biomrc \biolite have 2, 3, \dots, 20 candidate answers (top left), and the corresponding accuracy of the basic baselines (top right), and the neural models 
(bottom). \basethreeplus is the best basic baseline for 2 and 3 candidates, and for 2 candidates it is competitive to the neural models. Overall, however, \basefour is clearly the best basic baseline, but it is outperformed by all neural models in almost all cases, as in Table~\ref{tab:tab3}. \scibertmaxreader is again the best system in both settings, almost always outperforming the other systems. \asreader is the worst neural model in almost all cases. \aoareader is competitive to \scibertsumreader in Setting A, and slightly better overall than \scibertsumreader in Setting B, as can be seen in Table~\ref{tab:tab3}. 

\subsection{Results on \biomrc \biotiny}

\begin{figure*}[tbh]
\resizebox{\textwidth}{!}{
\begin{tabular}{|c|l|}
\hline
Passage & 
\begin{tabular}[c]{@{}p{15.2cm}@{}}
The study enrolled 53 \cyan{@entity1} (29 males, 24 females) with \orange{\textbf{@entity1576}} aged 15-88 years. Most of them were 59 years of age and younger. In 1/3 of the \cyan{@entity1} the diseases started with symptoms of \cyan{@entity1729}, in 2/3 of them--with pulmonary affection. \cyan{@entity55} was diagnosed in 50 \cyan{@entity1} (94.3\%), acute \cyan{@entity3617} --in 3 \cyan{@entity1}. ECG changes were registered in about half of the examinees who had no cardiac complaints. 25 of them had alterations in the end part of the ventricular ECG complex; rhythm and conduction disturbances occurred rarely. Mycoplasmosis \cyan{@entity1} suffering from \cyan{@entity741} ( \cyan{@entity741} ) had stable ECG changes while in those free of \cyan{@entity741} the changes were short. \cyan{@entity296} foci were absent. \cyan{@entity299} comparison in \cyan{@entity1} with \orange{\textbf{@entity1576}} and in other \cyan{@entity1729} has found that cardiovascular system suffers less in acute mycoplasmosis. These data are useful in differential diagnosis of \cyan{@entity296} .
\end{tabular}\\ \hline
Candidates &
\begin{tabular}[c]{@{}p{15.2cm}@{}}
\cyan{@entity1} : {[}`patients'{]} ;
\orange{\textbf{@entity1576}} : {[}`respiratory mycoplasmosis'{]} ;
\cyan{@entity1729} : {[}`acute respiratory infections', `acute respiratory viral infection'{]} ;
\cyan{@entity55} : {[}`Pneumonia'{]} ;
\cyan{@entity3617} : {[}`bronchitis'{]} ;
\cyan{@entity741} : {[}`IHD', `ischemic heart disease'{]} ;
\cyan{@entity296} : {[}`myocardial infections', `Myocardial necrosis'{]} ;
\cyan{@entity299} : {[}`Cardiac damage'{]} .
\end{tabular}\\ \hline
Question & 
\begin{tabular}[c]{@{}p{15.2cm}@{}}
Cardio-vascular system condition in \magenta{XXXX} .
\end{tabular}\\ \hline
Expert Human Answers & 
\begin{tabular}[c]{@{}p{15.2cm}@{}}
annotator1: \orange{\textbf{@entity1576}};
annotator2: \orange{\textbf{@entity1576}}.
\end{tabular}\\ \hline
Non-expert Human Answers & 
\begin{tabular}[c]{@{}p{15.2cm}@{}}
annotator1: \cyan{@entity296}; 
annotator2: \cyan{@entity296}; 
annotator3: \orange{\textbf{@entity1576}}.
\end{tabular}\\ \hline
Systems' Answers & 
\begin{tabular}[c]{@{}p{15.2cm}@{}}
\asreader: \cyan{@entity1729}; 
\aoareader: \cyan{@entity296}; 
\bertreaderf: \orange{\textbf{@entity1576}}.
\end{tabular}\\ \hline
\end{tabular}
}
\caption{
Example from \biomrc \biotiny. In Setting A, humans see both the pseudo-identifiers (@entity$N$) and the original names of the biomedical entities (shown in square brackets). Systems see only the pseudo-identifiers, but the pseudo-identifiers have global scope over all instances, which allows the systems, at least in principle, to learn entity properties from the entire training set. In Setting B, humans no longer see the original names of the entities, and systems see only the pseudo-identifiers with local scope (numbering reset per passage-question instance). 
}
\vspace*{-5mm}
\label{fig:fig2}
\label{fig:BioMRCExample_human}
\end{figure*}

\newcite{Pappas2018BioReadAN} asked humans (non-experts) to answer 30 questions from \bioread in Setting A, and 30 other questions in Setting B. We mirrored their experiment by providing 30 questions (from \biomrc \biolite) to three non-experts (graduate \textsc{cs} students) in Setting A, and 30 other questions in Setting B. We also showed the same questions of each setting to two biomedical experts. As in the experiment of \newcite{Pappas2018BioReadAN}, in Setting A both the experts and non-experts were also provided with the original names of the biomedical entities (entity names before replacing them with @entity$N$ pseudo-identifiers) to allow them to use prior knowledge; see the top three zones of Fig.~\ref{fig:fig2} for an example. By contrast, in Setting B the original names of the entities were hidden. 

Table \ref{tab:tinyresults} reports the human and system accuracy scores on \biomrc \biotiny. Both experts and non-experts perform better in Setting A, where they can use prior knowledge about the biomedical entities. The gap between experts and non-experts is three points larger in Setting B than in Setting A, presumably because experts can better deduce properties of the entities from the local context. Turning to the system scores, \scibertmaxreader is again the best system, but again much of its performance is due to the max-aggregation of the scores of multiple occurrences of entities. With sum-aggregation, \scibertsumreader obtains exactly the same scores as \aoareader, which again performs better than \asreader. (\aoareader and \scibertsumreader make different mistakes, but their scores just happen to be identical because of the small size of \biotiny.) Unlike our results on \biomrc \biolite, we now see all systems performing better in Setting A compared to Setting B, which suggests they do benefit from the global scope of entity identifiers. Also, \scibertmaxreader performs better than both experts and non-experts in Setting A, and better than non-experts in Setting B. However, \biomrc \biotiny contains only 30 instances in each setting, and hence the results of Table~\ref{tab:tinyresults} are less reliable than those from \biomrc \biolite (Table~\ref{tab:tab3}). 

In the corresponding experiments of \newcite{Pappas2018BioReadAN}, which were conducted in Setting B only, the average accuracy of the (non-expert) humans was 68.01\%, but the humans were also allowed not to answer (when clueless), and unanswered questions were excluded from accuracy. On average, they 
did not answer 21.11\% of the questions, hence their accuracy drops to 46.90\% if unanswered questions are counted as errors. In our experiment, the humans were also allowed not to answer (when clueless), but we counted unanswered questions as errors, which we believe better reflects human performance. Non-experts answered all questions in Setting A, and did not answer 13.33\% (4/30) of the questions on average in Setting B. 
The decrease in the questions non-experts did not answer (from 21.11\% to 13.33\%) in Setting B (the only one considered in \bioread) again suggests that the new dataset is less noisy, or at least that the task is more feasible for humans, even when the names of the entities are hidden. Experts did not answer 2.5\% (0.75/30) and 1.67\% (0.5/30) of the questions on average in Settings A and B, respectively.  

\begin{table}[tbh]
\begin{center}
\resizebox{\columnwidth}{!}{
\begin{tabular}{ccccccc}
\hline
\textbf{Method} & \textbf{Setting A} & \textbf{Setting B} \\
\hline \hline
\textbf{Experts (Avg)}      & \textbf{85.00}     & \textbf{61.67} \\
\textbf{Non-Experts (Avg)}  & 81.67     & 55.56 \\ 
\hline \hline
\textbf{\asreader}         & 66.67     & 46.67 \\
\textbf{\aoareader}         & 70.00     & 56.67 \\ 
\textbf{\bertreaderf}       & 70.00     & 56.67 \\
\textbf{\bertreaderfmax}    & \textbf{90.00}     & \textbf{60.00} \\
\hline
\end{tabular}
}
\vspace*{-2mm}
\caption{Accuracy (\%) on \biomrc \biotiny. Best human and system scores shown in bold. 
}
\vspace*{-5mm}
\label{tab:tinyresults}
\end{center}
\end{table}

Inter-annotator agreement was also higher for experts than non-experts in our experiment, in both Settings A and B (Table~\ref{tab:interannotator}). In Setting B, the agreement of non-experts was particularly low (47.22\%), possibly because without entity names they had to rely more on the text of the passage and question, which they had trouble understanding. By contrast, the agreement of experts was slightly higher in Setting B than Setting A, possibly because without prior knowledge about the entities, which may differ across experts, they had to rely to a larger extent on the particular text of the passage and question. 

\section{Related work}

Several biomedical \mrc datasets exist, but have orders of magnitude fewer questions than \biomrc \cite{Abacha_etal_2019_question_entailment_qa} or are not suitable for a cloze-style \mrc task \cite{pampari_etal_2018_emrqa, abacha_etal_2019_MEDIQA2019, Zhang_et_al_2018_Medical_Exam_QA}.
The closest dataset to ours is \clicr \cite{suster_et_al_2018_clicr}, a biomedical \mrc dataset with cloze-type questions created using full-text articles from \textsc{bmj} case reports.\footnote{\url{https://casereports.bmj.com/}} 
\clicr contains 100k passage-question instances, the same number as \biomrc \biolite, but much fewer than the 812.7k instances of \biomrc \biolarge. \v{S}uster et al.\ used \textsc{clamp} \cite{Soysal_et_al_2017_CLAMP} to detect biomedical entities and link them to concepts of the \textsc{umls} Metathesaurus \cite{Lindberg_et_al_1993_UMLS}. 
Cloze-style questions were created from the `learning points' (summaries of important information) of the reports, by replacing biomedical entities with placeholders. 
\v{S}uster et al.\ experimented with the Stanford Reader \cite{Chen2017ReadingWT} and the Gated-Attention Reader \cite{dhingra_etal_2017_GAReader}, which perform worse than \aoareader \cite{Cui2017AttentionoverAttentionNN}.

The \qa dataset of \bioasq \cite{Tsatsaronis_et_al_bioasq} 
contains questions written by biomedical experts.
The gold answers comprise multiple relevant documents per question, relevant snippets from the  documents, exact answers in the form of entities, as well as reference summaries, written by the experts. Creating data of this kind, however, requires significant expertise and time. In the eight years of \bioasq, only 3,243 questions and gold answers have been created. It would be particularly interesting to explore if larger automatically generated datasets like \biomrc and \clicr could be used to pre-train models, which could then be fine-tuned for human-generated \qa or \mrc datasets. 

\begin{table}[t]
\begin{center}
\setlength{\tabcolsep}{22pt}
\begin{tabular}{cc}
\hline
\textbf{Annotators (Setting)}         & \textbf{Kappa}           
\\ \hline \hline
\textbf{Experts (A)}       & 70.23     
\\
\textbf{Non Experts (A)}   & 65.61    
\\ \hline \hline
\textbf{Experts (B)}       & 72.30     
\\
\textbf{Non Experts (B)}   & 47.22   
\\ \hline \hline
\end{tabular}
\vspace*{-2mm}
\caption{Human agreement (Cohen's Kappa, \%) on \biomrc \biotiny. Avg.\ pairwise scores for non-experts.}
\vspace{-7mm}
\label{tab:interannotator}
\end{center}
\end{table}

Outside the biomedical domain, several cloze-style open-domain \mrc datasets have been created automatically \cite{Hill_et_al_CBTest, Moritz_Hermann_et_al_cnndaily, Dunn_etal_2017_SearchQA, Bajgar_et_al_BookTest}, but have been criticized of containing questions that can be answered by simple heuristics like our basic baselines \cite{Chen_et_al_vs_cnndailymail}. There are also several large open-domain \mrc datasets annotated by humans  \cite{Kwiatkowski_et_al_Natural_Questions, Rajpurkar2016SQuAD10, rajpurkar_etal_2018_squadv2, Trischler_etal_2016_NewsQA, Nguyen_et_al_msmarco, lai_etal_2017_RACE_dataset}.
To our knowledge the biggest human annotated corpus is Google's Natural Questions dataset \cite{Kwiatkowski_et_al_Natural_Questions}, with approximately 300k human annotated examples.
Datasets of this kind require extensive annotation effort, which for open-domain datasets is usually crowd-sourced. Crowd-sourcing, however, is much more difficult for biomedical datasets, because of the required expertise of the annotators.  

\section{Conclusions and Future Work}

We introduced \biomrc, a large-scale cloze-style biomedical \mrc dataset.
Care was taken to reduce noise, compared to the previous \bioread dataset of \newcite{Pappas2018BioReadAN}. Experiments showed that \biomrc's questions cannot be answered well by simple heuristics, and that two neural \mrc models that had been tested on \bioread perform much better on \biomrc, indicating that the new dataset is indeed less noisy or at least that its task is more feasible. Human performance was also higher on a sample of \biomrc compared to 
\bioread, and biomedical experts performed even better. We also developed a new \bert-based model, the best version of which outperformed all other methods tested,  reaching or surpassing the accuracy of biomedical experts in some experiments. We make \biomrc available in three different sizes, also releasing our code, and providing a leaderboard. 

We plan to tune more extensively the \bert-based model to further improve its efficiency, and to investigate if some of its techniques (mostly its max-aggregation, but also using sub-tokens) can also benefit the other neural models we considered. We also plan to experiment with other \mrc models that recently performed particularly well on open-domain \mrc datasets \cite{Zhang_et_al_bestinsquad_albert}. 
Finally, we aim to explore if pre-training neural models on \bioread is beneficial in human-generated biomedical datasets \cite{Tsatsaronis_et_al_bioasq}. 

\section*{Acknowledgments}

We are most grateful to I.~Almirantis, S.~Kotitsas, V.~Kougia, A.~Nentidis, S.~Xenouleas, who participated in the human evaluation with \biomrc \biotiny.

\label{main:ref}

\bibliographystyle{acl_natbib}
\bibliography{biomrc}

\end{document}